\documentclass{article}
\usepackage{ijcai25}

\usepackage{times}
\usepackage{soul}
\usepackage{url}
\usepackage[hidelinks]{hyperref}
\usepackage[utf8]{inputenc}
\usepackage[small]{caption}
\usepackage{graphicx}
\usepackage{amsmath}
\usepackage{amsthm}
\usepackage{booktabs}
\usepackage{algorithm}
\usepackage{algorithmic}
\usepackage[switch]{lineno}
\usepackage{subfigure}

\title{Knowledge Guided Encoder-Decoder Framework: Integrating Multiple Physical Models for Agricultural Ecosystem Modeling}

\author{
    Qi Cheng\textsuperscript{1} \and
    Licheng Liu\textsuperscript{2} \and
    Yao Zhang\textsuperscript{3} \and
    Mu Hong\textsuperscript{3} \and
    Shiyuan Luo\textsuperscript{1} \and
    Zhenong Jin\textsuperscript{2} \and
    Yiqun Xie\textsuperscript{4} \and
    Xiaowei Jia\textsuperscript{1}
    \affiliations
    \textsuperscript{1}University of Pittsburgh \and
    \textsuperscript{2}University of Minnesota - Twin Cities \and
    \textsuperscript{3}Colorado State University \and
    \textsuperscript{4}University of Maryland, College Park
    \emails
    \textsuperscript{1}\{qic69, shl298, xiaowei\}@pitt.edu,
    \textsuperscript{2}\{lichengl, jinzn\}@umn.edu,
    \textsuperscript{3}\{yao.zhang, mu.hong\}@colostate.edu,
    \textsuperscript{4}xie@umd.edu
}

\usepackage{multirow}
\usepackage{booktabs}
\usepackage{booktabs}
\usepackage{colortbl}
\usepackage{amsmath}
\usepackage[table]{xcolor}

\begin{document}

\maketitle

\begin{abstract}
Agricultural monitoring is critical for ensuring food security, maintaining sustainable farming practices, informing policies on mitigating food shortage, and managing greenhouse gas emissions. Traditional process-based physical models are often designed and implemented for specific situations, and their parameters could also be highly uncertain. In contrast, data-driven models often use black-box structures and does not explicitly model the inter-dependence between different ecological variables. As a result, they require extensive training data and lack generalizability to different tasks with data distribution shifts and inconsistent observed variables. To address the need for more universal models, we propose a knowledge-guided encoder-decoder model, which can predict key crop variables by leveraging knowledge of underlying processes from multiple physical models. 
The proposed method also integrates a language model to process complex and inconsistent inputs and also utilizes it to implement a model selection mechanism for selectively combining the knowledge from different physical models. Our evaluations on predicting carbon and nitrogen fluxes for multiple sites demonstrate the effectiveness and robustness of the proposed model under various scenarios.

\end{abstract}

\section{Introduction}
Agricultural ecosystems are facing increasing pressures due to the growing population and the changing climate. Effective modeling of agricultural ecosystems can provide critical and timely insights into the growth of crops, soil health, water conditions, and impacts of environmental changes. These insights are important for  
ensuring food security, maintaining sustainable farming practices, and creating policies to mitigate natural disturbance-incurred food shortage (e.g., distribution of subsidies). Moreover, such models can help provide  information needed for managing greenhouse gas (GHG) emissions.  According to the statistics from the U.S. Department of Agriculture (USDA), around 10\% of GHG emissions  are related to agriculture management in the United States~\cite{USDA2021}. Hence, the effective monitoring of agricultural ecosystems is indispensable for making decisions and policies for mitigating agricultural emissions and hence climate change. 

Conventional process-based physical models (PBMs) have been widely used to simulate GHG emissions from agricultural land. While grounded in detailed scientific principles, these physical models often involve complex parameterization~\cite{parton2018daycent,zhou2021quantifying,markstrom2012p2s}. Such parameterization approximates the reality under certain conditions and observations and thus lacks flexibility in adapting to real-world scenarios with varying environmental conditions,  management practices, and modeling scales. 
Moreover, multiple PBMs have been developed for distinct tasks using different datasets. As a result,  different PBMs may show advantageous performance in modeling different processes in agricultural ecosystems. 
Machine learning (ML)-based data-driven models, on the other hand, are capable of learning patterns automatically from large data. Hence, they are increasingly considered as alternatives to PBMs for scientific modeling~\cite{reichstein2019deep,xu2015data,o2018using,willard2022integrating}.  However, traditional ML models are not designed to capture complex physical and biochemical processes in scientific systems. Moreover, training these models usually requires large training data, which is often unavailable in real agricultural monitoring tasks due to the substantial human labor and material costs needed for data collection. Recent works have shown the potential of enhancing ML  with scientific knowledge in agricultural modeling~\cite{liu2024knowledge,liu2022kgml}. Despite their promise, these methods require extensive expertise to design models for a specific task and are not easily adaptable to other tasks.    

To address these challenges, this paper proposes a knowledge-guided foundation model (KGFM) that leverages complementary strengths of PBMs and ML models.  
Such novel combination of physics and ML within the KGFM method provides threefold benefits. First, it can better capture complex physical processes by using a modular encoder structure that replicates different components in the agricultural ecosystems, including the carbon cycle, water cycle, energy cycle, and nitrogen cycle. This encoder is trained with sufficient simulated data produced by PBMs to capture dynamics specific to each component. It also enhances the interpretability by providing the simulation of flux variables related to intermediate processes. 
Second, it embraces multiple PBMs and introduces a model selection mechanism to intelligently identify PBMs that are most relevant to the downstream task. Hence, this significantly reduces the need for domain expertise for common users or stakeholders to manually select suitable models. 
The knowledge of the relevant PBMs is then leveraged to enhance ML model prediction. 
Finally, KGFM is able to simulate a range of physical variables related to the agricultural ecosystems through a decoder structure. 
In particular, it leverages language model-based structures to enable flexibility in handling partial observations so that KGFM can still perform model selection and get fine-tuned when we just observe certain variables (e.g., CO$_2$ or N$_2$O) in a downstream task (e.g., for a specific monitoring location).



We evaluate the proposed KGFM model using true observations of GPP, N$_2$O, and CO$_2$ from different regions. The results demonstrate the superiority of KGFM in predicting different target variables compared to other baseline methods. We also validate the effectiveness of the model selection mechanism in identifying relevant PBMs. 

Our implementation has been released\footnote{\url{https://anonymous.4open.science/r/KGFM-7264/}}, and the dataset will also be made publicly available upon acceptance. 
This work represents a significant step toward more robust, generalizable, and interpretable modeling for scientific systems, paving the way for future research 
that integrates accumulated  knowledge into ML for advancing scientific discovery.  
\section{Related Work}


The proposed method can be generally applied to modeling environmental ecosystems in many domains, such as climate science, hydrology, and geology. 
Traditional scientific modeling often relies on PBMs that are built based on a series of mathematical or physical equations to simulate the different processes. These models are necessarily approximations of reality, and their parameterization components make them limited in adapting to different scenarios, as highlighted by prior studies~\cite{osborne2010modeling,wang2023parameterization}. Here we introduce recent works that use ML models as an alternative to modeling environmental ecosystems. 


\textbf{ML methods for modeling environmental ecosystems. } ML-based data-driven approaches have gained popularity due to their ability to handle large datasets and uncover complex patterns without requiring explicit process understanding. ML models, such as neural networks and random forests, have been applied to predict carbon fluxes and other ecological variables, as demonstrated by studies from Carbonneau et al. \cite{carbonneau2020machine} and Sun et al. \cite{sun2022ecological}. More recently, advanced models, such as graph neural networks and transformer-based models, have shown further performance improvement in certain environmental  applications~\cite{topp2023stream,ye2021transformer}. 
However, these methods are generally limited in their interpretability and generalizability, especially given limited and less representative training data 
\cite{schneider2017ecological,read2019process}.

\textbf{Knowledge Guided ML Models. } Knowledge guided ML (KGML) has emerged as a new hybrid paradigm to leverage complementary strengths of ML models and physical models~\cite{willard2022integrating,karpatne2024knowledge}. In particular, KGML aims to incorporate physical knowledge into different components of ML models, including training objective, model architecture, and learning algorithms.  For example, previous works have shown that using the unsupervised mass and energy conservation loss can help improve the prediction of water quantity~\cite{frame2023strictly}, water quality~\cite{jia2021physics_tds}, and crop yield~\cite{he2023physics}, especially given limited training data. The knowledge from existing PBMs can also be transferred to ML models. For example, Liu et al. introduced a  KGML model that simulates intermediate processes using the process-based Ecosys model to enhance the carbon cycle quantification in agroecosystems \cite{liu2024knowledge}. 
Additionally, the physical knowledge can be used in pre-training~\cite{han2021transfer,read2019process,jia2021physics_simlr}  or model adaptation~\cite{chen2023physics}. In summary, these works have demonstrated the promise of KGML in capturing complex patterns and generalizing to unseen scenarios, especially given limited training data.


\textbf{Foundation models for environmental science. } 
Foundation models offer new opportunities for scientific modeling given their ability to harness large and complex data and the flexibility in handling different input and output structures. For example, existing works have shown that large language models (LLMs) can be used to extract semantic embeddings from meteorological data, which can be better understood by language models for modeling environmental ecosystems~\cite{luo2023free,li2024lite}. It is also shown that LLMs can be used to handle inconsistent data and incorporate additional information in the prompt~\cite{luo2023free}. Additionally, there is a growing interest in building foundation models for complex temporal dynamics~\cite{nguyen2023climax,bi2022pangu}, which is known to be a challenging problem for standard data-driven models.  However, many existing foundation models developed for text and vision domains are not exposed to data specific to the scientific problems and thus may not perform well when adapted to the scientific modeling task with limited observations. These foundation models still rely on pure data-driven methods but largely ignore accumulated knowledge in scientific domains. 
\section{Proposed Method}

\begin{figure*} [!h]
\centering
\subfigure[]{ \label{fig:a}{}
\includegraphics[width=0.43\linewidth]{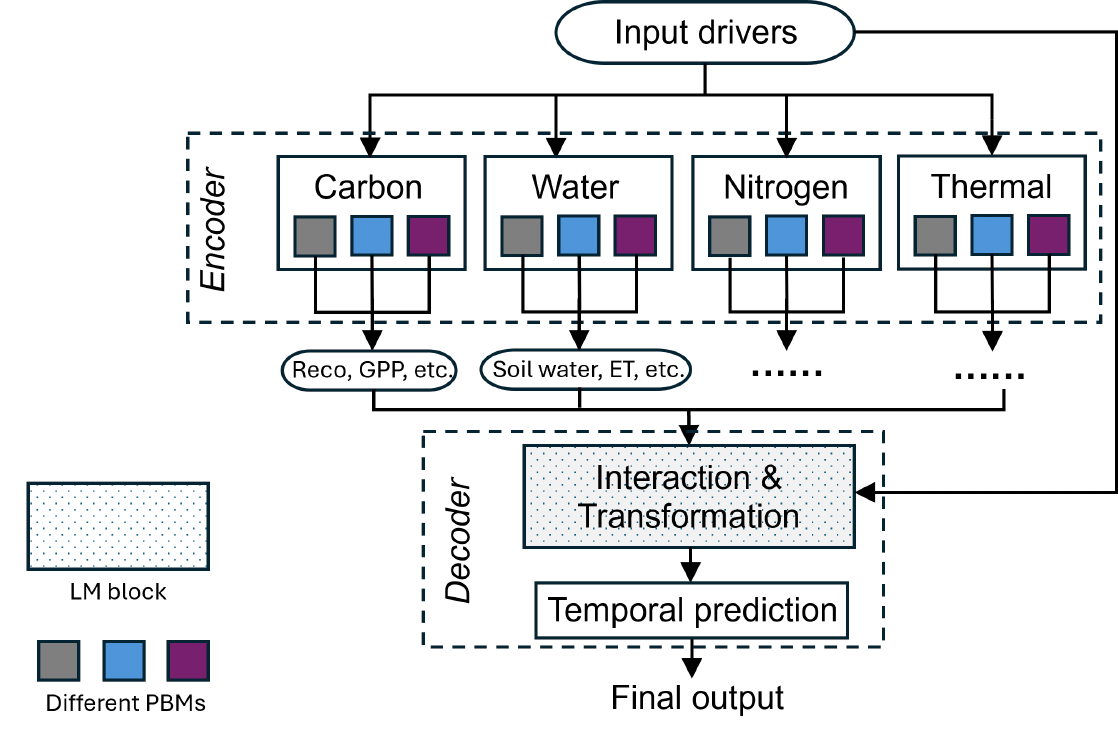}
} \hspace{.3in}
\subfigure[]{ \label{fig:b}{}
\includegraphics[width=0.48\linewidth]{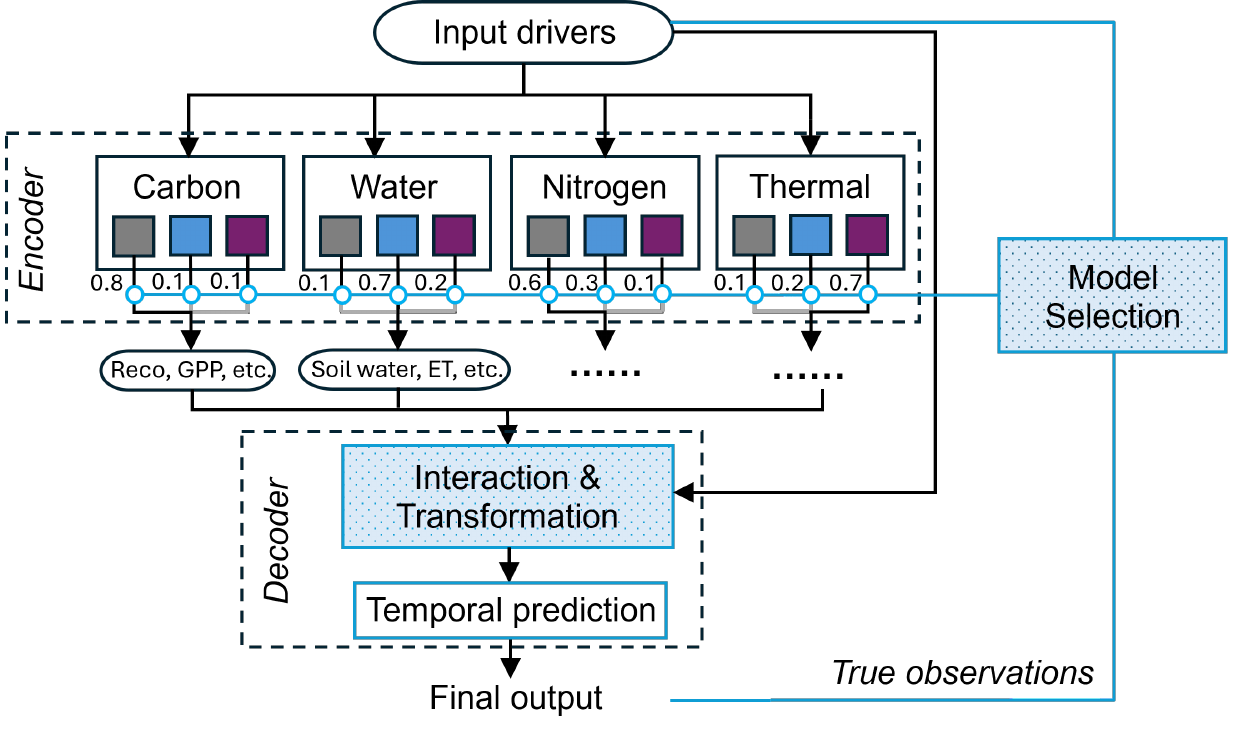}
}
\vspace{-0.1in}
\caption{(a) The overall encoder-docoder architecture of KGFM. The encoder extracts intermediate flux variables, which are combined with input drivers in the decoder to predict final outputs. (b) Adaptation of KGFM to the downstream task. Different PBM surrogates are weighted and the decoder is fine-tuned given true observations. 
}
\vspace{-0.1in}
\label{fig:steps}
\end{figure*}

The KGFM method integrates knowledge from PBMs into an encoder-decoder architecture to model agricultural ecosystems. The encoder leverages the knowledge from multiple PBMs to create a modular structure, which separately simulates intermediate processes related to 
carbon, nitrogen, water, and thermal cycles. More formally, given the input drivers $\mathbf{x}=\{\mathbf{x}^1, \mathbf{x}^2,...,\mathbf{x}^T\}$ over $T$ dates in each location, where each $\mathbf{x}^t$  includes 16 variables related to weather, soil, and management information,  the encoder extracts intermediate flux variables $\mathbf{v}=\{\mathbf{v}^1, \mathbf{v}^2,...,\mathbf{v}^T\}$. 
The decoder then combines the intermediate flux variables $\mathbf{v}$  and produces the final target variables $\hat{\mathbf{y}}=\{\hat{\mathbf{y}}^1, \hat{\mathbf{y}}^2,...,\hat{\mathbf{y}}^T\}$. This process is shown in Fig.~\ref{fig:steps} (a). Given the input drivers $\mathbf{x}$ and observation data $\mathbf{y}$ from a downstream task,  KGFM utilizes an additional model selection structure to identify PBMs that have advantageous modules relevant to the downstream task and increase their weights in making prediction, as shown in Fig.~\ref{fig:steps} (b).


The following subsections detail the key components of the KGML framework and their implementation.

\subsection{Knowledge-guided Modular Encoder}

PBMs are often built with a molecular structure to mimic the separable yet interactive carbon, nitrogen, water, and thermal cycles that jointly determine the physical and biogeochemical changes of the agricultural system. 
Existing ML models largely ignore the modular structure and try to directly build a mapping to all target variables. We build an encoder structure that employs the modular structure (i.e., carbon, water, nitrogen, and thermal cycles), in which the processes for different modules are modeled separately and do not interfere with each other. Module interactions will be considered later using the decoder.  

Incorporation of PBMs enables transferring physical knowledge into the ML model, and thus can potentially enhance the model generalizability~\cite{willard2022integrating}. However, selecting the appropriate PBM can be challenging because different PBMs may have varying performance in downstream tasks, depending on their origin, focus, and complexity. To address this issue, we propose to include multiple PBMs in the encoder to perform comprehensive feature extraction. 

However, the incorporation of PBMs into the ML model is challenging as they require expensive computational costs in inference due to   complex numerical solvers embedded in these models. 
It is therefore infeasible to directly incorporate PBMs into KGFM, which needs to be trained over large data. Hence, we  build data-driven surrogates for each PBM. 
The effectiveness of these surrogate models depends on their ability to accurately emulate the physical processes within each  module, which has been supported by the results presented in  Table 1 of the supplement. Moreover,  data-driven surrogates offers enhanced flexibility, which allows  selective integration of modules from different physical models. For example, one could combine a carbon  surrogate from one physical model with a nitrogen surrogate from another physical model. 

In particular, for the $n^{th}$ PBM, we create surrogate models separately for different modules (i.e., carbon, nitrogen, water, and thermal). 
Each surrogate model is developed using a two-layer Gated Recurrent Unit (GRU) architecture. The first layer takes  input drivers over $T$ time steps, i.e., $\mathbf{x}=\{\mathbf{x}^1,\mathbf{x}^2,..., \mathbf{x}^T\}$, 
to predict intermediate ecosystem variables (e.g., soil water and bulk density), as 
$\mathbf{q}_n=\text{GRU}_{n,1}(\mathbf{x})$. The second layer combines the original  input features with the output from the first layer to produce the flux variables related to each module (e.g., CO$_2$ flux for the carbon module and evapotranspiration for the water module), as 
\begin{equation}
{\mathbf{v}_n}=\text{GRU}_{n,2}([\mathbf{x},\mathbf{q}_n]). 
\end{equation}

The encoding process is illustrated in Fig.~\ref{fig:steps} (a). Then we combine the output flux variables from $N$ different modules, i.e., $\mathbf{v}=[\mathbf{v}_1,..., \mathbf{v}_N]$, and feed them to the decoder. 

The PBM surrogates in the encoder are trained using  
synthetic data generated by PBMs under different scenarios. Such simulated data can provide rich information about agricultural ecosystems under variation of environmental conditions and management practices, while we often have only limited observation data for certain scenarios. %



Another challenge is that different PBMs may be designed to output different flux variables from each module. Such inconsistency could bring complexity for building a unified decoder to process the intermediate flux variables from multiple modules. This issue will be addressed later when we discuss the design of the decoder.





\subsection{Language Model (LM)-enhanced Decoder}

We built an LM-enhanced decoder for processing  intermediate flux variables, handling module interactions, and generating target output variables. 
The choice of the LM-enhanced decoder is also motivated by the aforementioned key challenge in incorporating physical models: handling inconsistent variables across different PBMs and observational datasets. 
In particular, the LM-enhanced decoder consists of two major blocks, as shown in Fig.~\ref{fig:steps}~(a). The first block leverages a language model to gather the intermediate flux variables generated by all the modules and embeds these variables on each date in a semantic space. The second block then employs a temporal network to process the daily embeddings over long time periods and make the final prediction.

\subsubsection{LM for feature embedding }  The LM block aims to leverage the power of language models to capture complex interactions and dependencies within the intermediate flux variables 
and the forcing data. 
Specifically, 
the obtained variables need to be first converted to a format that can be accepted by language models. In this work, we employ the linearization process~\cite{wang2023meditab,luo2023free}, which combines variable names and their values, separated by commas, as
\begin{equation}
\text{linearize}(\mathbf{v}^t)= \{{c}_1:\mathbf{v}^t_{(1)}, {c}_2:\mathbf{v}^t_{(2)}, ..., {c}_K:\mathbf{v}^t_{(K)}\}, 
\end{equation}
where $c_k$ and $\mathbf{v}^t_{(k)}$ denote the name and  value for the $k^{th}$ variable in $\mathbf{v}^t$, respectively. 
For example, a simple linearized output can be  ``$CO_2$ 
 flux: 13, GPP: 25, Reco: -8, N$_2$O flux: -0.0.003''. 


We then utilize a pre-trained BERT model to embed the obtained intermediate flux variables and the original input forcing data at each time step $t$, as $\mathbf{u}^t = \text{BERT}(\text{linearize}([\mathbf{v}^t_\text{carbon},\mathbf{v}^t_\text{water},\mathbf{v}^t_\text{nitrogen},\mathbf{v}^t_\text{thermal},\mathbf{x}^t]))$, where $\mathbf{v}_{*}^t$ represents the flux variables for a module within the encoder, i.e., $*\in \{\text{carbon},\text{water},\text{nitrogen},\text{thermal}\}$, at time $t$. The BERT model we use is a pretrained DistilBERT base model uncased \cite{Sanh2019DistilBERTAD}.

In addition, the LM block offers an advantage in managing inconsistent flux variables. For instance, LMs like BERT do not require input sentences to have fixed lengths or rigid structures, and thus can effectively process linearized sequences that include varying sets of variables, e.g., one model could produce GPP,  CO$_2$, and Reco as carbon fluxes while the other model just produces GPP and CO$_2$. 

\subsubsection{Temporal prediction layers } 
The embeddings $\{\mathbf{u}^t\}_{t=1}^T$ are first fed to a long-short term memory (LSTM) to encode long-term temporal dependencies over time, as $\{\mathbf{h}^t\}_{t=1}^T = \text{LSTM}(\{\mathbf{u}\}_{t=1}^T)$. Then the representation $\mathbf{h}^t$ is passed through linear layers to make prediction of target variables.  


While the BERT has been pre-trained with large data, it has not seen adequate linearized data about environmental ecosystems. Hence, we also tune both the LM layer and the temporal prediction layers using simulated data. 
These decoder layers aim to capture the general relationship from intermediate fluxes and input drivers to the final target, and thus are trained using a combination of simulated data generated by different PBMs.


\subsection{Model Selection and Fine-tuning}

We now discuss the model adaptation process to a downstream task with small observed data samples, as illustrated in Fig.~\ref{fig:steps} (b). 
We first introduce a model selection layer to identify  PBM surrogates that are relevant to the downstream task. The importance of these relevant PBM surrogates is then increased in making final prediction. Additionally, the decoder layers are fine-tuned towards new observation data from the downstream task. In the following, we detail the model selection mechanism.



\begin{figure}[ht]
\centering
\includegraphics[width=0.49\textwidth]{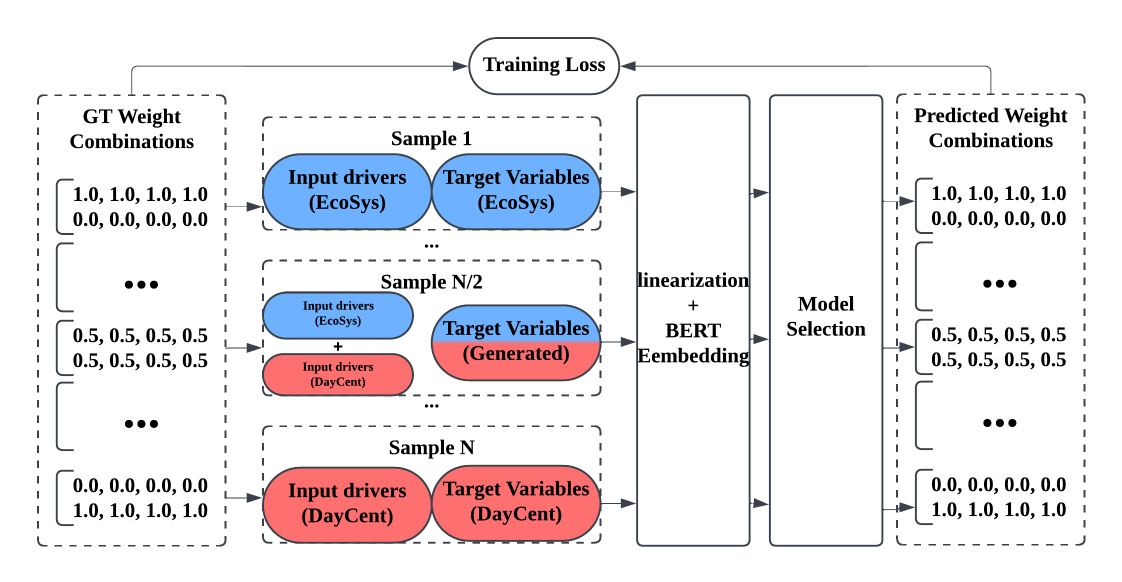}
\caption{Training Process of the model selection module.}
\label{fig:ms_traning}
\end{figure}

\subsubsection{Assigning weights to PBM surrogates }
The proposed KGFM employs a fine-grained model ensemble approach, which adaptively reweights PBM surrogates in each module (i.e., carbon, water, nitrogen, and thermal) and combines them for a  downstream task. 
This adaptive weighting mechanism addresses a common challenge in agricultural monitoring and many other environmental studies.  Multiple PBMs are often built for similar problems while their internal structures and parameters are set to fit datasets of very different locations, spatial scales, time periods, and variables. For example, DayCent has been optimized with extensive US agricultural data for national greenhouse gas inventories, while Ecosys~\cite{Grant2001,Zhou2021} has been optimized on North American cropping systems. Consequently, it can be extremely difficult for  clients or stakeholders to manually select relevant PBMs for their own downstream tasks, as this requires substantial domain expertise ~\cite{Necpalova2015,DelGrosso2020}. The proposed KGFM addresses this issue by automatically identifying relevant PBMs and leveraging their complementary strengths in improving the prediction.  


In particular, the model selection mechanism aims to assign weights $\{\alpha_{m,n}\}_{m=1,n=1}^{M,N}$ for $M$ different PBM surrogates in $N$ different modules ($N=4$ in our study). Ideally, the weight $\alpha_{m,n}$ should reflect the relevance of the m$^{th}$ PBM surrogate for the $n^{th}$ module, with higher value indicating that this PBM  is more suitable for the $n^{th}$ module in the downstream task. 
Once the weights are determined, KGFM computes the weighted combination of the outputs of the PBM surrogates, as the output of each encoder module, and then feeds it into the decoder for making predictions. 


Now we describe the process of computing the weights $\{\alpha_{m,n}\}$. Given new observations, i.e., pairs of input drivers $\mathbf{x}^t$ and observed target variables $\mathbf{y}^t$, we conduct the linearization over the combination of $[\mathbf{x}^t, \mathbf{y}^t]$, and then feed the output to a pre-trained BERT model to create the embedding, as $\mathbf{z}^t = \text{BERT}(\text{linearize}([\mathbf{x}^t, \mathbf{y}^t]))$. It is noteworthy that this can be the same BERT model as the one used in the decoder.  Moreover, the observations could contain only a subset of target variables as LM can process linearized sequences with different elements. 

Given that new observations are often available on multiple dates, we compute the weights for each time step from $\mathbf{z}^t$ using fully connected layers and then take the average over all the steps. More formally, the predicted weights are computed as follows: 
\begin{equation}
\{\alpha_{m,n}\} = \text{Avg} \{\text{MLP}(\mathbf{z}^t)\}_\mathcal{O},
\end{equation}
where $\text{Avg}$ denotes the averaging  operation, $\text{MLP}$ denotes the multi-layer fully connected network, and $\mathcal{O}$ represents all the observations in the downstream task. 
The output dimension of MLP is $M\times N$, which is normalized over every $M$ PBM surrogates within each module $n$.


\subsubsection{Training procedure }
Since  KGFM involves non-differentiable operations, such as data-to-text conversion (i.e., linearization), the model selection component (e.g., parameters in MLP) cannot be directly tuned through back-propagation from predicted errors on final target variables. 
Instead, we directly train the model selection component using simulated data. The idea is to create synthetic observations of target variables by combining different PBM simulations under certain preset weights, and then tuning the model selection component to predict the preset weights given the input of synthetic observations.  

To ensure the model selection component is trained over a comprehensive set of scenarios, we exhaustively preset PBM weights and create synthetic observations. 
For example, if there are two PBMs in each of the four modules, we will consider weight variation for the first PBM  
from $[0.1, 0.1, 0.1, 0.1]$ to $ [0.9, 0.9, 0.9, 0.9]$ for four different modules with an interval of 0.1 for each dimension. Hence in total, we consider 9$^4$ combinations. This can also be easily extended to consider a larger number of PBMs.   


\section{Evaluation}
\subsection{Dataset}


To develop and validate the proposed KGFM, we use both simulated and observational datasets. 
The simulated datasets are generated by PBMs and are designed to simulate realistic agroecosystem dynamics by capturing the complex interactions between various environmental, soil, and management factors. The observation datasets were employed for downstream tasks, focusing on predicting agricultural carbon and nitrogen outcomes. 
Although KGFM can predict a wide range of agricultural ecosystem variables, we focus our evaluation on three key variables: GPP, CO$_2$, and N$_2$O, which represent key carbon and nitrogen fluxes. Other variables are often less available due to the difficulties in data collection. 

\begin{table*}[t]
\small
\centering
\caption{Coefficient of determination ($R^2$) of various models for output feature GPP, CO$_2$, and N$_2$O on observation datasets. KGFM is the proposed framework. The keyword inside the parenthesis indicates which PBM surrogates are used in the encoder while MS refers to the model selection mechanism. FT means the model is fine-tuned on observation datasets, otherwise the models (in the top block) are  trained  with only simulated data. The best and the second best models are indicated by bold and asterisk ($*$), respectively.}
\begin{tabular}{c|llllll}
\hline
                                            & \multicolumn{2}{c}{GPP}         & \multicolumn{2}{c}{CO$_2$}      & \multicolumn{2}{c}{N$_2$O}      \\ \hline
\multicolumn{1}{l|}{}                       & $R^2$          & $RMSE$         & $R^2$          & $RMSE$         & $R^2$          & $RMSE$         \\ \cline{2-7} 
LSTM                                        & 0.454          & 0.980          & 0.225          & 1.141          & 0.300          & 1.816          \\
PG-AN \cite{he2023physics} & 0.072          & 1.278          & -0.384         & 1.525          & -0.069         & 2.245          \\
Transformer                                 & 0.408          & 1.021          & 0.073          & 1.248          & 0.070          & 2.094          \\
KGFM (Ecosys)                               & 0.486          & 0.951          & 0.308          & 1.079          & 0.207          & 1.933          \\
KGFM (DayCent)                              & 0.513          & 0.926          & 0.523          & 0.895          & 0.340          & 1.764          \\
KGFM (MS)                                   & 0.492          & 0.946          & 0.501          & 0.916          & 0.317          & 1.794          \\ \hline
LSTM FT                                     & 0.759          & 0.651          & 0.549          & 0.871          & 0.469          & 1.582          \\
PG-AN FT                                    & 0.037          & 1.302          & 0.525          & 0.894          & 0.178          & 1.968          \\
Transformer FT                              & 0.550          & 0.890          & 0.359          & 1.038          & 0.061          & 2.104          \\
KGFM (Ecosys) FT                            & 0.770          & 0.636          & \textbf{0.661} & \textbf{0.755} & 0.554          & 1.450          \\
KGFM (DayCent) FT                           & 0.795$^*$      & 0.601$^*$      & 0.608          & 0.812          & \textbf{0.625} & \textbf{1.329} \\
KGFM (MS) FT                                & \textbf{0.806} & \textbf{0.584} & 0.649$^*$      & 0.768$^*$      & 0.613$^*$      & 1.351$^*$      \\ \hline
\end{tabular}
\label{tab:real-data-performance}
\vspace{-.15in}
\end{table*}

\subsubsection{Simulated data. }

The simulated datasets were generated using PBMs, Ecosys and Daycent, which simulate agroecosystem dynamics under various scenarios. For the Ecosys model, we generated daily synthetic data for the period 2000–2018 across 99 randomly selected counties in Iowa, Illinois, and Indiana states of USA. To introduce variability, we implemented 20 different N fertilization rates ranging from 0 to 33.6 g N m$^{-2}$ in each county. For the DayCent model, simulations were conducted daily for 2,562 sites randomly sampled in the US Midwest from 2000 to 2020. Each site was modeled under 42 scenarios, varying by N fertilizer rates from 0 to 33.6 g N m$^{-2}$, fertilization timing (at planting or 30 days after), and crop rotation (corn-soybean or soybean-corn). These models are parameterized to reflect real-world conditions, providing a valuable testing ground for the KGML framework's capabilities. More details about these  PBMs are included in Appendix. By leveraging synthetic data, the study ensures comprehensive validation of the framework's predictive performance before deploying it on actual field data, thus minimizing potential biases or noise that may arise from incomplete observational datasets.

 \begin{figure*}[ht]
\centering
\includegraphics[width=0.95\textwidth]{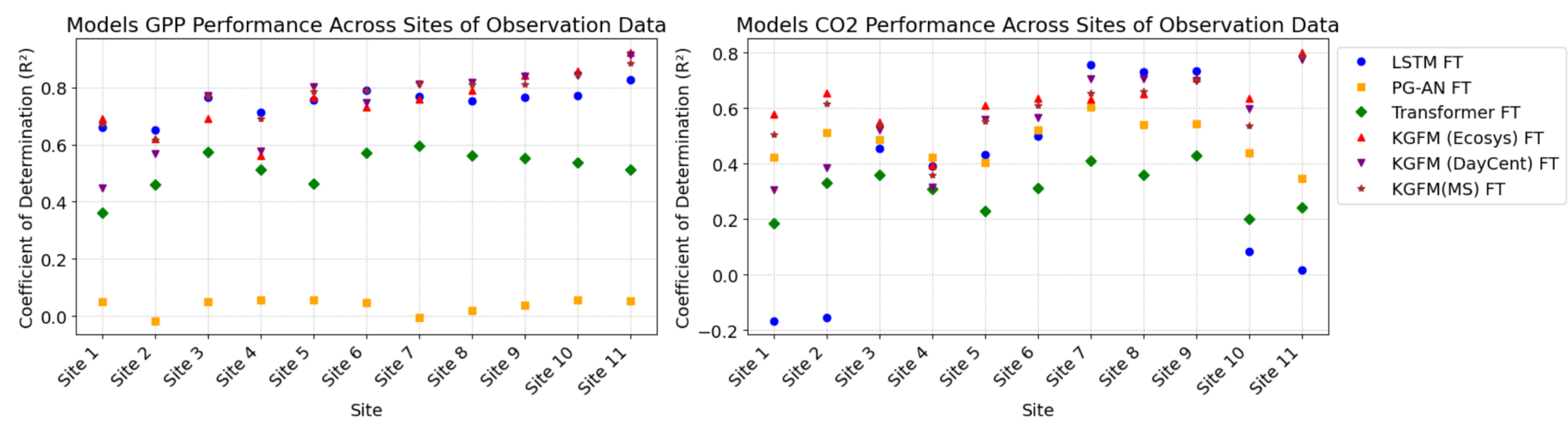}
\vspace{-.1in}
\caption{Model performance for predicting GPP and CO$_2$ across sites with observation data. }
\label{fig:gpp_site}
\vspace{-.15in}
\end{figure*}

\subsubsection{Observation dataset (N$_2$O)} was collected from a controlled-environment facility 
using soil samples from a corn-soybean rotation farm~\cite{gmd-15-2839-2022}. Six chambers were used to grow continuous corn during 2016-2018, with precise monitoring of N$_2$O fluxes in response to different precipitation treatments from April 1st to July 31st. N$_2$O fluxes were measured hourly and processed for a daily time scale, alongside soil moisture, nitrate, ammonium concentrations, and environmental variables. Simulating agricultural N$_2$O emissions is important for mitigating climate change but challenging due to its hot moment/spot and underlying complex biogeochemical processes. This dataset is crucial for evaluating KGFM's ability to capture these N$_2$O dynamics. 

\subsubsection{Observation dataset (GPP and CO$_2$)} was derived from 11 cropland eddy covariance (EC) flux tower sites located in major U.S. corn and soybean production regions~\cite{a4eb67e9e8a9432c9674ec9dcb78f494}. These sites, including US-Bo1, US-Bo2, US-Br1, US-Br3, US-IB1, US-KL1, US-Ne1, US-Ne2, US-Ne3, US-Ro1, and US-Ro5, span across Illinois, Iowa, Michigan, Nebraska, and Minnesota states. The GPP data was decomposed from observed CO$_2$ fluxes at these sites using the ONEFlux tool. The weather data were retrieved from the EC flux towers, while soil information and plant type information were retrieved from gSSURGO, and CDL data. The dataset provided daily time scale measurements, covering a time span from 2000 to 2020 across 11 cropland EC flux tower sites, with each site having different operational periods, ranging from 5 to 19 years. The dataset provides comprehensive temporal and spatial variance for validating KGFM's ability to capture agricultural carbon flux dynamics. 

\subsubsection{Data Integration and Preprocessing. }

All datasets underwent a series of preprocessing steps to ensure compatibility and quality. The input features were normalized to facilitate the model's learning process, while missing values were imputed using domain-specific techniques to maintain data integrity. The datasets were then divided into training and testing subsets, ensuring an unbiased evaluation of the model's performance. 

The simulated dataset consists of three dimensions: days of the year, sites, and features, and are used to train the PBM surrogates. 
For the observation dataset, we split the training set and testing set by the dimension years instead of sites so that we can observe performance on all the sites and chambers. {For the GPP and CO$_2$ dataset, we use the data from 2000-2015 for training and the data from 2016-2020 for testing. For the N$_2$O observation dataset, we use the data from 2016-2017 for training and the data from 2018 for testing. This temporal split strategy prevents any future information from leaking into the training process.}




\subsection{Predictive Performance}
\begin{figure*}[ht]
\centering
\includegraphics[width=0.98\textwidth]{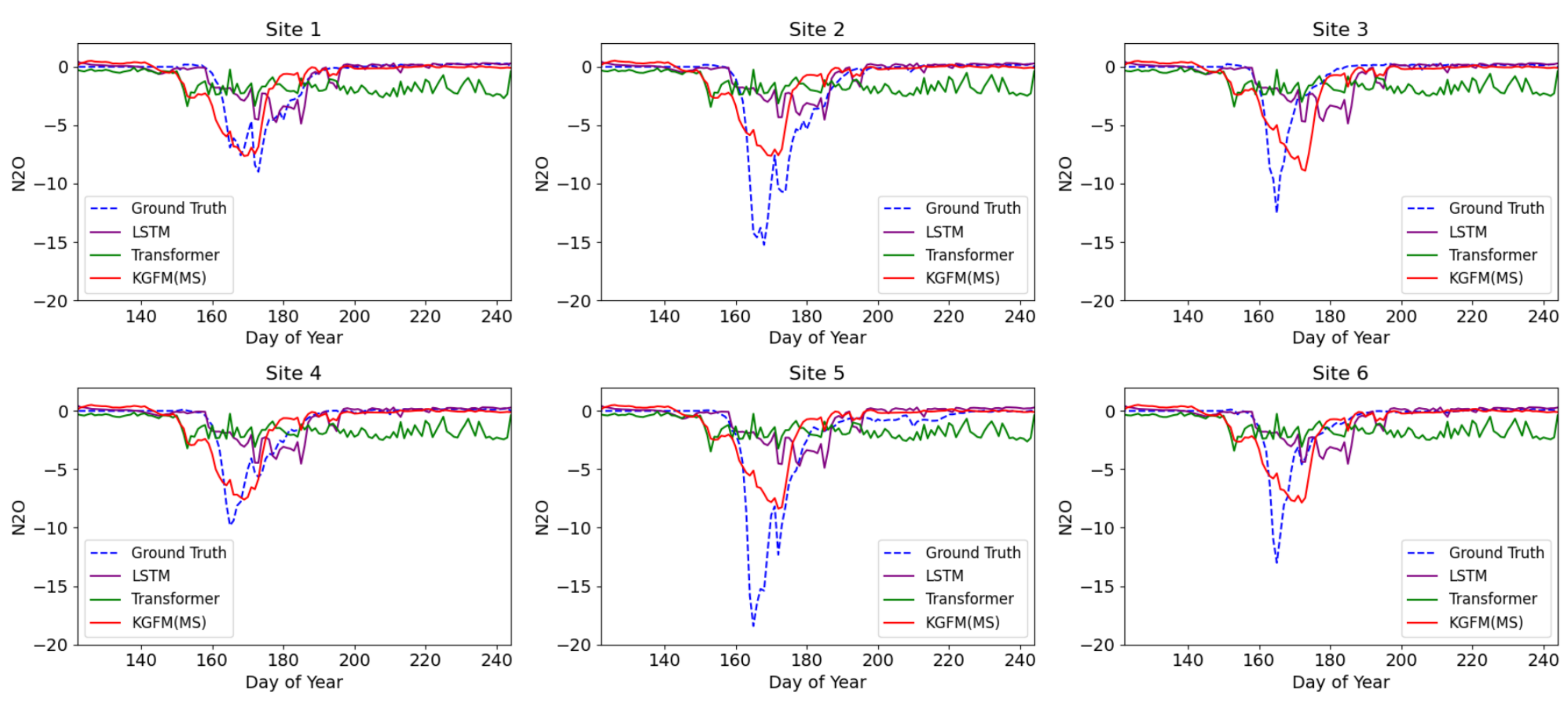}
\vspace{-.1in}
\caption{Predictions of scaled N$_2$O values for a specific year across six different sites {in the testing set of the observation data}. Each subplot represents one site, with the x-axis indicating the day of the year and the y-axis showing the scaled N$_2$O values. From this plot, it is observed proposed KGFM model consistently outperforms the other models, particularly in accurately capturing the peak N$_2$O values, which are the most critical for N$_2$O predictions.}
\label{fig:n2o_ts}
\vspace{-.1in}
\end{figure*}

\begin{figure*}[!htbp]
\centering
\includegraphics[width=0.98\textwidth]{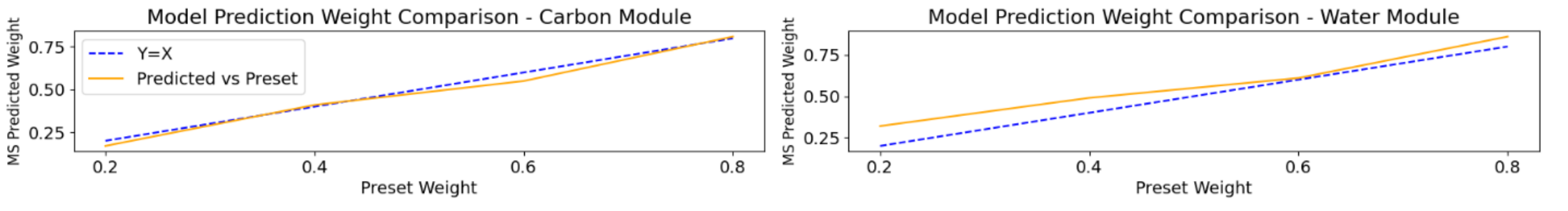}
\vspace{-.1in}
\caption{Comparison between predicted and ground truth weights for the two encoder modules: carbon, and water. Each subplot presents the weight variation of the Ecosys for the specific module being tested, while the weights for the other three modules remain constant across the four combinations. The figure for experiments on nitrogen and thermal module are in Appendix}
\label{fig:ms_eval}
\vspace{-.1in}
\end{figure*}

Table~\ref{tab:real-data-performance} presents the testing performance of KGFM and multiple baselines. Each method is implemented in two configurations: (1) training exclusively with simulated data, and (2) fine-tuning the pre-trained model using real observations in the downstream task. The baselines include a standard LSTM module, the PG-AN model \cite{he2023physics}, and standard transformer modules. In the first configuration, all baselines and KGFM variations were trained on a combined training set from two simulated datasets for 40 epochs using a learning rate of 0.001, and then tested on the observation datasets' testing set. In the second configuration, the pre-trained models were further fine-tuned on the training set of the observation data for 10 epochs with a learning rate of 0.0001. The Adam optimizer and mean squared error (MSE) loss function were used for all training processes. Model training was performed on an RTX 3090 GPU, with training involving language models using two A1000 GPUs.

We can observe that the KGFM-based methods generally perform much better than other baselines in each scenario from Table~\ref{tab:real-data-performance}. The LSTM performs better than the Transformer model because it can better capture long-term temporal dependencies, which are critical for modeling the dynamics of agricultural ecosystems.  

When using only simulated data for training, KGML (MS) performs slightly worse than KGFM (DayCent) but better than KGML (Ecosys), as shown in the top block of Table
~\ref{tab:real-data-performance}. It can be observed that some the surrogates learned from DayCent perform much better than those learned from Ecosys when they are directly evaluated on true observations in our downstream tasks. KGFM (MS) is able to automatically find a balance between them even without being fine-tuned. 

After fine-tuning, all the methods have better performance. Similar observation can be made that  KGFM (MS) FT  achieves performance comparable to or even better than the higher-performing model from KGFM (Ecosys) FT and KGFM (DayCent) FT. We further show the performance at each site of GPP and CO$_2$ in Figure~\ref{fig:gpp_site}. 
We also include the site-wise performance for N$_2$O in Appendix. These results confirm that KGFM with the model selection mechanism can achieve the performance that is comparable to or even better than existing models even without prior knowledge of which PBMs are best suited for the downstream task.


\subsection{Tracking Temporal Changes}
An essential objective for modeling environmental ecosystems is to effectively capture temporal changes, which is crucial for informed decision-making in management practices. While some methods, such as LSTM, can achieve reasonable overall performance, they tend to fit general trends rather than precisely tracking specific changes.
In Figure~\ref{fig:n2o_ts}, we show the  N$_2$O fluxes predicted by different methods on different observation sites. The KGFM method demonstrates superior performance than LSTM and Transformer in capturing temporal changes in N$_2$O fluxes. This improvement can be attributed to the integration of PBMs within different physical modules, which enables the simulation of intermediate flux variables, and consequently, better informing the dynamics of target variables. We also validate the effectiveness of KGFM in emulating PBMs for estimating intermediate flux variables and final target variables in Appendix.

\subsection{Evaluation of the Model Selection Mechanism}
To validate the model selection mechanism, we create synthetic data that combines the outputs of Ecosys and DayCent PBMs with preset weights. Then KGFM performs model selection using the synthetic data. 
Figure~\ref{fig:ms_eval} shows the relation between 
the predicted weights and the preset weights. The results confirm that KGFM can effectively adjust the weights of PBM surrogates according to different datasets fed to the model selection component.

\section{Conclusion}
In this paper, we proposed KGFM as a general paradigm for modeling target ecosystems by leveraging accumulated scientific knowledge from PBMs. The knowledge guidance facilitates capturing complex processes and learning with limited observation data. 
KGFM's model selection mechanism also enables automatically selecting and combining  PBMs 
, which facilitates the use of KGFM by stakeholders. Besides, the KGFM is able to produce intermediate flux variables and thus can provide more information beyond initial target variables for tracking the ecosystem dynamics. Our evaluations have demonstrated the effectiveness of KGFM in predicting several key variables (e.g., CO$_2$, N$_2$O, GPP). 

The proposed KGFM serves as a stepping stone for building general foundation models for scientific discovery. Given that real scientific problems often involve complex dynamics and patterns that are not yet fully understood, it is critical to build models upon accumulated scientific knowledge. 
In addition, KGFM enhances both accuracy and efficiency compared to traditional PBMs, which enables a broader adoption of advanced agricultural monitoring and simulation. This could benefit clients with varying levels of expertise and supports regions with limited observational data.
Future work will include incorporating more PBMs and established theories.

\newpage
\bibliographystyle{named}
\bibliography{reference,Xiaowei,Qi}
\end{document}